\documentclass{article}

\usepackage{arxiv}

\usepackage[utf8]{inputenc} 
\usepackage[T1]{fontenc}    
\usepackage{hyperref}       
\usepackage{url}            
\usepackage{booktabs}       
\usepackage{amsfonts}       
\usepackage{amsmath}        
\usepackage{nicefrac}       
\usepackage{microtype}      
\usepackage{cleveref}       
\usepackage{graphicx}
\usepackage{natbib}
\usepackage{doi}
\usepackage{capt-of}        
\usepackage{authblk}        

\title{Predicting Consequences and Reinforcing Navigation Policies with Latent World Models}

\date{}

\newbox{\orcid}\sbox{\orcid}{\includegraphics[scale=0.06]{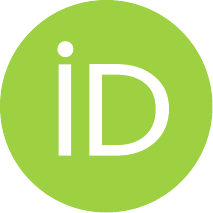}}
\author[1,2]{\href{https://orcid.org/0009-0000-7029-6546}{\usebox{\orcid}\hspace{1mm}Zengmao Wang}}
\author[2,3]{\href{https://orcid.org/0000-0003-2257-5684}{\usebox{\orcid}\hspace{1mm}Wei Gao}\thanks{Corresponding author.}}
\author[2,3]{\href{https://orcid.org/0000-0002-8704-7914}{\usebox{\orcid}\hspace{1mm}Shuhan Shen}}
\affil[1]{School of Advanced Interdisciplinary Sciences, University of Chinese Academy of Sciences}
\affil[2]{Institute of Automation, Chinese Academy of Sciences}
\affil[3]{School of Artificial Intelligence, University of Chinese Academy of Sciences}

\renewcommand{\shorttitle}{Latent World Models for Navigation Policy Reinforcement}

\hypersetup{
pdftitle={Predicting Consequences and Reinforcing Navigation Policies with Latent World Models},
pdfsubject={},
pdfauthor={Zengmao Wang, Wei Gao, Shuhan Shen},
pdfkeywords={World model, Robot learning, Visual navigation},
}

\begin{document}
\maketitle

\begin{abstract}
  World models enable agents to reason about future outcomes and learn policies from their knowledge of state transition, but existing approaches primarily focus on reconstructing future observations or features, which introduces unnecessary complexity and limits their effectiveness for decision making. In this work, we propose a compatibility prediction Latent World Model (LWM) for robot navigation that predicts action-conditioned latent feature compatibility rather than reconstructing observations. Our key insight is that spatial proximity correlates with latent feature similarity, enabling action consequences to be evaluated directly in latent space. To support counterfactual training, our model leverages action sequences sampled across trajectories and learns to predict which sequences lead closer to the goal. Furthermore, we demonstrate how the learned world model can supervise policy learning from unlabeled video data and further improve policies through reinforcement learning entirely within the world model. This imagination-driven framework eliminates the need for action annotations and additional environment interaction. Extensive experiments on multiple real-world robot navigation datasets show that our approach significantly outperforms prior world model and imitation learning methods in prediction accuracy, policy learning, and real-world navigation performance. The code, pretrained models, and additional materials are available at \href{https://wzm206.github.io/latent-world-model-nav/}{https://wzm206.github.io/latent-world-model-nav/}.
\end{abstract}


\section{Introduction}
\label{sec:intro}

Autonomous navigation requires robots to reason about the consequences of their actions before executing them in the real world. World models have emerged as a promising paradigm for enabling such predictive reasoning by learning environment dynamics from offline data \cite{ha2018recurrent}. By simulating future outcomes internally, world models allow agents to plan \cite{bar2025navigation}, evaluate \cite{li2025worldeval}, and improve policies \cite{ye2026world} without costly real-world interaction. This paradigm is particularly attractive for robotics, where data collection is expensive and safety constraints limit online exploration.

However, existing world models suffer from a fundamental limitation: they are typically trained to reconstruct future observations conditioned strictly on the factual actions executed in the dataset \cite{bar2025navigation, hafner2019dream}. When we use the world model for planning or decision-making, the model needs to infer situations that have not actually occurred, also known as counterfactual reasoning. This creates a critical misalignment: while planning requires counterfactual reasoning to evaluate alternative actions, training relies solely on ground-truth trajectories. Furthermore, optimizing for pixel or latent reconstruction introduces unnecessary complexity, compounding errors, and representations that are misaligned with decision-making. We argue that world models should focus on predicting action consequences rather than generating exact observations.

In this work, we propose a fundamentally different perspective: instead of predicting future observations, we train a world model to predict the compatibility between action consequences and future latent state which leads counterfactual reasoning into the training phase. Our approach is motivated by a simple yet powerful observation: in navigation tasks, spatial proximity strongly correlates with visual feature similarity. Therefore, rather than reconstructing exact future observations, it is sufficient to predict the compatibility.

To achieve this counterfactual training, we introduce an imagination-based world model that leverages action sequences from different trajectories to simulate counterfactual futures in latent space. Given a current observation, our model predicts latent state features resulting from multiple candidate action sequences and learns to distinguish which sequences lead to states closer to the ground-truth future. Crucially, simulating these counterfactual scenarios during training significantly maximizes data utilization efficiency because collecting real-world interaction data is notoriously expensive and challenging in robotics field. This formulation enables the model to learn robust, decision-centric representations without requiring pixel-level prediction. Therefore, we achieve training world models in counterfactual imagination.

Beyond world model learning, we demonstrate how the learned model can be used to train and improve navigation policies from unlabeled video data. We first use the world model to label candidate action sequences according to their predicted similarity to goal states, enabling supervised policy learning without ground-truth actions. We then further improve the policy using reinforcement learning entirely within the learned world model. A well-trained world model can naturally serve as a reward model. This imagination-driven reinforcement learning enables policy improvement without additional environment interaction.

We evaluate our approach in real environments and three robot navigation datasets. Our method demonstrates significant improvements over prior world model and navigation methods. Our results show that compatibility prediction provides a powerful and efficient alternative to generative world modeling, enabling effective planning and policy learning from offline data. We summarize our contributions as follows.
\begin{itemize}

\item A new world model formulation based on latent compatibility prediction.
We propose to predict action-conditioned latent feature compatibility rather than reconstructing future observations, enabling efficient and decision-oriented world modeling.

\item A counterfactual imagination framework using cross-trajectory action sequences.
Our method leverages action sequences from different trajectories to simulate alternative futures, enabling counterfactual reasoning from purely offline data.

\item A unified framework for policy learning and reinforcement entirely within the world model. We show how unlabeled video data can be used to supervise policy learning and further improve policies via imagination-based reinforcement learning.

\item State-of-the-art performance in real-world environment and robot navigation benchmarks.

\end{itemize}

\section{Related Work}
Driven by advancements in large language models \cite{vaswani2017attention}, video generation \cite{ho2022video}, and embodied intelligence \cite{driess2023palm}, world models have garnered significant attention \cite{ding2025understanding}. While the concept continues to evolve, it primarily serves two functions: learning internal representations to understand the environment, and utilizing predictive modeling to guide decision-making for external actions.

\subsection{Building a World Model}
Numerous efforts focus on constructing world models. Video generation-based approaches \cite{wan2025wan, agarwal2025cosmos, kong2024hunyuanvideo, gao2025seedance} are intuitive and demonstrate remarkable physical consistency---such as understanding gravity and collisions---highlighting their potential as implicit world models \cite{hu2023gaia,du2023learning,ren2025videoworld,kang2024far}. These methods typically condition future frame prediction on current observations and actions. For instance, LingBot-World \cite{team2026advancing} employs a virtual rendering pipeline for interactive data generation, NWM \cite{bar2025navigation} leverages large-scale real-world datasets to generate diverse action-conditioned videos, and 3D-VLA \cite{zhen20243d} extends this paradigm to 3D point clouds.

However, pixel-level video generation suffers from slow inference speeds and inconsistent physics. Dino-WM \cite{zhou2024dino} addresses this by shifting to feature generation using a Dino-v2 encoder \cite{oquab2023dinov2}, though its primary objective remains reconstruction. In contrast, our method avoids pixel-space generation entirely, preventing the model from fixating on irrelevant details. By optimizing to predict the consequences of actions in the latent space rather than reconstructing features, our approach eliminates the need for iterative diffusion, enabling rapid prediction and decision-making.

\subsection{Learning in the World Model}
Ultimately, world models aim to facilitate decision-making and learning for agents. Learning paradigms within these models generally fall into two categories.

\subsubsection{Joint Training of World and Policy Models}
Akin to model-based reinforcement learning \cite{ha2018world, ha2018recurrent, hafner2019learning, silver2017predictron}, these methods jointly optimize both models. PlaNet \cite{hafner2019learning} introduces the RSSM architecture to enable dynamic learning in a compact latent space. Dreamer \cite{hafner2019dream} and DreamerV2 \cite{hafner2020mastering} learn latent dynamics from images and optimize policies via imagined trajectories, excelling in long-horizon tasks. PWM \cite{georgiev2024pwm} facilitates first-order policy optimization within regularized world models. However, these methods often require continuous agent-environment interaction, confining them to simulated environments and limiting real-world practicality. While our method also leverages latent state representations, it can be deployed directly in the real world without requiring online interaction.

\subsubsection{Utilizing Frozen World Models as Simulators}
Alternatively, frozen world models can serve as realistic simulators. DreamZero \cite{ye2026world} learns physical dynamics via joint video-action prediction for real-time control. World-env \cite{xiao2025world} replaces physical interactions with a virtual simulator for RL training. RWM \cite{li2025robotic} uses a dual-autoregressive model for robust sim-to-real transfer, and WorldVLA \cite{cen2025worldvla} unifies VLA and world models. These approaches typically train policies on synthesized pixel-level images, which is significantly more complex than predicting feature-space similarities. Furthermore, the generalization of video generation remains problematic \cite{kang2024far}. For instance, autonomous driving models struggle to imagine out-of-distribution scenarios like driving on sidewalks due to a lack of training data. Conversely, our cross-trajectory imagination approach can effectively evaluate such deviations, keeping the agent safely aligned with the goal.

\begin{figure}[htb]
  \centering
  \includegraphics[height=6.2cm]{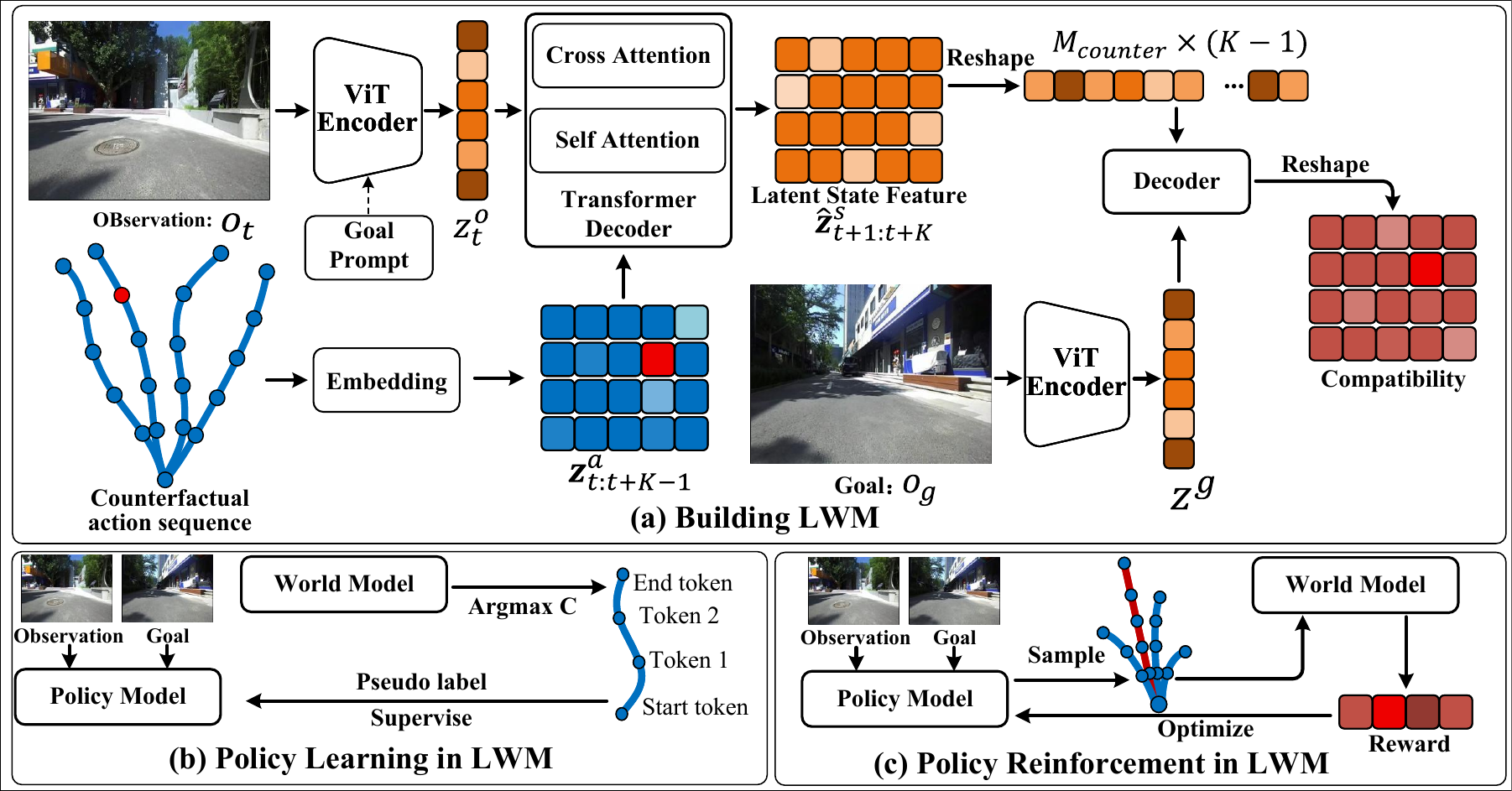}
  \caption{The overview of LWM. (a) When training the world model, we need to predict the consequences of different action sequences and obtain compatibility with the goal. (b) LWM can provide pseudo labels for the policy model. (c) The policy model can be reinforced in LWM.
  }
  \label{fig:example}
\end{figure}

\section{Method}

We propose a compatibility prediction Latent World Model (LWM) that learns action-conditioned latent dynamics by predicting feature compatibility rather than reconstructing future observations. As shown in Fig. \ref{fig:example}, our framework enables counterfactual reasoning from offline trajectories to alleviate the challenges of data annotation and collection. Our framework supports policy learning and reinforcement entirely within LMW. We will elaborate on our method in detail below.

\subsection{General Problem Formulation}

We consider a dataset of offline trajectories:

\begin{equation}
\mathcal{D} = \{ \tau_i \}_{i=1}^N, \quad
\tau_i = \{(o_1, a_1), (o_2, a_2), \dots, (o_T, a_T)\}
\end{equation}
where $o_t \in \mathbb{R}^{H \times W \times 3}$ is the visual observation and $a_t \in \mathbb{R}^{d_a}$ is the robot action at timestep $t$. In visual navigation tasks, ${d_a}=3$ represents the coordinates of motion $( x,y,\theta )$.

Our goal is to learn a world model $\text{WM}$ that can evaluate the consequences of arbitrary action sequences starting from a given observation. Within the world model, it can be divided into two modules: the prediction module and the discrimination module. In the prediction module, the world model predicts the state features after executing the action. In the discrimination module, the world model determines the compatibility between the predicted features and the goal.
\begin{equation}
\label{eq:wm}
\hat{\mathbf{z}}_{t+1:t+K}=\text{WM}_\text{predict}(o_t, A_{t:t+K-1}), \quad {C}_{t+1:t+K}=\text{WM}_\text{discriminative}(\hat{\mathbf{z}}_{t+1:t+K}, o_{g})
\end{equation}
where $o_t$ is the observation at time $t$, $A_{t:t+K-1} = \{a_t, a_{t+1}, \dots, a_{t+K-1}\}$ is a future action sequence, $\hat{\mathbf{z}}_{t+1:t+K}$ is the predicted features and $C$ is the compatibility.

In video generation world models, $\hat{\mathbf{z}}_{t+1:t+K}$ in Eq.(\ref{eq:wm}) could be considered as images generated based on observation and action sequences. $C$ in Eq.(\ref{eq:wm}) is obtained through pre-trained feature loss, such as comparing the LPIPS \cite{zhang2018unreasonable} loss between the generated image and the goal image. Unlike these works, we do not attempt to reconstruct future observations. Instead, we bring the entire process into the latent space for better decision-making and planning.

\subsection{Latent World Model}
\subsubsection{Latent Observation and Action Encoding}

We first map observations into a latent feature space using a visual encoder:

\begin{equation}
\label{eq:encoder_obs}
z_t^o = E_o(o_t), \quad z_t \in \mathbb{R}^{N_p \times D}
\end{equation}
where $E_o$ is a Vision Transformer \cite{dosovitskiy2020image} encoder. $N_p$ represents the number of patches in the images. Similarly, we encode each action into a latent token:

\begin{equation}
\mathbf{z}_{t:t+K-1}^a = E_a(A_{t:t+K-1}), \quad \mathbf{z}_{t:t+K-1}^a \in \mathbb{R}^{(K-1) \times D}
\end{equation}

Our world model predicts latent state features conditioned on the current state and an action sequence. We implement the world model prediction using a standard causal Transformer decoder:

\begin{equation}
\hat{\mathbf{z}}^s_{t+1:t+K} = \text{Decoder}(z_t^o, \mathbf{z}_{t:t+K-1}^a)
\end{equation}
Each prediction follows autoregressive causal masking. For each action token, it can only see its previous actions. At this point, we can predict $(K-1)$ future features. For previous methods, they often predicted pixels or features and optimized neural networks using reconstruction loss: $Loss=MSE(\hat{\mathbf{z}}, \mathbf{z})$. However, they did not take into account the assistance of other action sequences in the dataset. Although we don't know what the observations of performing other actions look like, we can estimate their compatibility with the goal images.

\subsubsection{Compatibility Learning Objective}
A key limitation of offline data is that each observation is paired with only one action sequence. To enable counterfactual reasoning, we construct alternative action sequences by sampling from other trajectories. Given a state $z_t^o$, we sample $M_{\text{counter}}$ candidate action sequences:

\begin{equation}
\mathcal{A}=\{A^{(1)}, A^{(2)}, \dots, A^{(M_{\text{counter}})}\}
\end{equation}
The world model predicts corresponding future latent state features:

\begin{equation}
\hat{\mathcal{Z}} = \{\hat {\mathbf{z}}^{s(1)}, \hat {\mathbf{z}}^{s(2)}, \dots, \hat {\mathbf{z}}^{s(M_{\text{counter}})}\}, \quad \hat {\mathbf{z}}_{t+1:t+K}^{s(i)}=\text{Decoder}(z^o, E_a(A^{(i)}))
\end{equation}

This allows the model to learn action-conditioned compatibility across diverse trajectories, enabling counterfactual reasoning without environment interaction. Our method relies on a simple and reasonable assumption: features $\mathbf{z}$ with closer spatial distances often more similar in the latent space. For simplicity, we consider training with only one random goal image $z^g = E_o(o_g)$ in ground truth action sequence $\{a_t, a_{t+1}, \dots, a_{t+K_{gt}}\}$. It is worth noting that the selection of goal image during training is random and $K_{gt}$ is smaller than $K$. Therefore, for each $\hat {z}^{s(i)}_j$, we can calculate a compatibility with the ground truth goal image. For $M_{\text{counter}}$ action sequences, we can obtain $M_{\text{counter}} \times (K-1)$ compatibility score:

\begin{equation}
    C = \left \{ S(\hat {z}^{s(i)}_j, z^g)  \right \} ^ {i=1,2,...,M_{\text{counter}}} _ {j=1,2,...,K-1}, \quad C \in \mathbb{R}^{M_{\text{counter}} \times (K-1)}
\end{equation}
where S is a neural network used for compatibility prediction. Because the state features are in the latent space, we still choose standard Transformer Decoder with a linear output layer instead of LPIPS \cite{zhang2018unreasonable}. Among $C_{ij}$, the score corresponding to ground truth action sequence $\{a_t, a_{t+1}, \dots, a_{t+K_{gt}}\}$ should be the highest. This naturally reminds us of classification objectives similar to contrastive learning. Specifically, we can consider one result $C_{gt}$ obtained from ground truth action sequences as positive samples and $M_{\text{counter}} \times (K-1) -1$ results obtained from counterfactual action sequences as negative samples. In this way, we can use Info-NCE loss \cite{gutmann2010noise} to optimize the world model.

\begin{equation}
    \mathcal{L}=-\log \frac{\exp C_{gt}}{ {\textstyle \sum_{i=1}^{M_{\text{counter}}} \sum_{j=1}^{K-1} \exp C_{ij} } }
\end{equation}
However, we found that although this approach can obtain a world model, it ignores the spatial relationships between action sequences. To better capture the underlying environment dynamics, we define the compatibility label in Eq. (\ref{eq:log}) using a logarithmic mapping of spatial distances. Physically, this transformation converts the geometric Euclidean metric into a high-fidelity compatibility density. The $-\log$ operator ensures that the supervision signal is highly sensitive to small spatial deviations, thereby forcing the world model to distinguish subtle differences between the ground-truth trajectory and its spatially-close neighbors (i.e., hard negatives).
\begin{equation}
    \label{eq:log}
    C_{\text{label}} = \{ - \log[{ d(a^i_j, a_{gt}) + \epsilon}]  \}^ {i=1,2,...,M_{\text{counter}}} _ {j=1,2,...,K-1}
\end{equation}
where $d$ represents the metric distance of space, and we choose the simplest Euclidean distance. $\epsilon$ is a small constant. In the spatial distance, the smallest element value is 0, which corresponds to the ground truth action. It is worth noting that this method can be directly used to train multiple goal images together with almost no increase in computational. We only need to replace $z_g$ and $a_{gt}$ with multiple ground truth on the same action sequence during training. Afterwards, we use simple MSE loss function: $\mathcal{L} = MSE(C, C_{\text{label}})$.

\subsubsection{Best practices}
\label{best_practice}
When we need to use world models for planning or decision-making, we first give the current observation $o_t$, different action sequences $\mathcal{A}$, and predict the executed features $\hat{\mathcal{Z}}$. Then we can get the compatibility score $C$ with the goal image we want to reach, and select the action sequence with the highest compatibility score to execute. In fact, this is still the paradigm of predicting first and then discriminating. This reminds us of a question: Can we use the information of the goal when encoding the image from the beginning? If the goal features are given first during prediction in Eq. (\ref{eq:encoder_obs}), we can get better results. That is to say, if the world model is for planning, we can provide goal information in advance when encoding observations and then determine the compatibility score. Then Eq. (\ref{eq:encoder_obs}) becomes $z_t^o = E_o(o_t, o_g)$. This approach is equivalent to giving a goal in advance, and then the world model imagines the compatibility score to the goal after performing different actions. This goal can not only be an image target, but can also be extended to semantic targets \cite{nie2025wmnav} and language instructions \cite{yao2025navmorph}. However, previous world model mainly focused on dynamic models of the environment without providing goal information in advance. Therefore, for the best performance and fair comparison, we test two options separately: providing goal prompts in advance and not providing them. The specific details can be further discussed in the experimental section.

\subsection{Policy Learning from World Model}
\label{policy_learning}

Once the world model is trained, it can be leveraged to utilize large-scale unlabeled video data for policy learning. We define a goal-conditioned policy $\pi(a_t \mid o_t, o_g)$ that takes the current observation $o_t$ and the goal observation $o_g$ as inputs, and outputs an action sequence that drives the robot toward the goal state.

Following prior autoregressive policy formulations such as OpenVLA \cite{kim2024openvla} and ArMoNa \cite{wang2025causal}, our policy generates actions as a sequence of discrete tokens. Specifically, we first compute action increments $\Delta a$ and apply k-means clustering to obtain $K_{\text{center}} = 64$ cluster centers, which serve as discrete action tokens. Any continuous action sequence can thus be represented as a sequence of these tokens. Our policy model consists of a visual encoder and a causal transformer decoder. The visual encoder extracts features from the current and the goal observation, while the causal decoder autoregressively predicts the action tokens. Similar to next-token prediction in natural language processing, the model generates tokens sequentially, starting from a special start token and continuing until an end token is produced.

In the absence of ground-truth action labels, we exploit the predictive capability of the world model to generate pseudo action labels from unlabeled video sequences $\{o_t, o_{t+1}, \dots\}$. We first construct a candidate action set by clustering existing action sequences into $M_c = 64$ cluster centers, denoted as $\mathcal{A}_c = \{A_c^{(1)}, A_c^{(2)}, \dots, A_c^{(M_c)}\}$. We then randomly sample observation-goal pairs from video trajectories. For each pair $(o_t, o_g)$, we evaluate all candidate action sequences using the compatibility score from world model. The action sequence with the highest compatibility score is selected as the pseudo label. The discretized action sequence can be represented as $\{a_{start}, a_t, a_{t+1}, \dots, a_{t+K_{\text{pre}}}, a_{end} \}.$

Finally, the policy model is trained using a standard autoregressive cross-entropy loss:
\begin{equation}
\mathcal{L}
=
-\sum_{i=t}^{t+K_{\text{pre}}}
\log
\pi_{\theta}
\left(
a_i \mid o_t, o_g, a_{<i}
\right),
\end{equation}
which encourages the model to generate action sequences that lead to the desired goal observation.

\subsection{Policy Reinforcement in Latent World Model}

After the initial imitation stage, the policy model acquires the basic ability to generate action sequences conditioned on the current and goal observations. However, this stage is limited by the fixed candidate action set constructed from clustering, which restricts policy expressiveness. To further improve the policy, we perform reinforcement learning entirely within the latent world model. This process is analogous to the cold-start reinforcement paradigm used in DeepSeek-R1 \cite{guo2025deepseek}, where the model is progressively improved beyond the initial imitation distribution.

Unlike the imitation stage, where the policy is trained on a fixed set of clustered action sequences, the autoregressive policy can now freely generate novel action sequences by composing discrete tokens. Given a current observation $o_t$ and a goal observation $o_g$, we sample multiple candidate action sequences from the pretrained policy:
\begin{equation}
\mathcal{A}=\{A^{(1)}, A^{(2)}, \dots, A^{(M_{\text{gen}})}\},
\quad A^{(i)} \sim \pi_{\theta}(\cdot \mid o_t, o_g).
\end{equation}

After predicting the consequences of different actions, the discrimination module of LWM can naturally be seen as a reward model. Specifically, for each sampled action sequence $A^{(i)}$, the world model evaluates its compatibility with the goal observation, producing a scalar reward:
\begin{equation}
\hat{z}^i=\text{WM}_\text{predict}(o_t, A^{(i)}), \quad r_i=\text{WM}_\text{discriminative}(\hat{z}^i, o_{g})
\end{equation}
where higher reward indicates that the predicted outcome is closer to the goal state. To stabilize training, we normalize the rewards across sampled sequences to obtain advantages:
\begin{equation}
A_i = \frac{r_i - \text{mean}(\mathbf{r})}{\text{std}(\mathbf{r})}.
\end{equation}
We optimize the policy using a GRPO-style \cite{guo2025deepseek} objective, which encourages high-reward actions while constraining the policy to remain close to a reference policy:
\begin{align}
\mathcal{L}_{\text{GRPO}}
&=
-\frac{1}{M_{\text{gen}}}
\sum_{i=1}^{M_{\text{gen}}}
\left(
r_{\text{ratio}} A_i
-
\beta \mathbb{D}_{KL}
\left(
\pi_{\theta}
\;\|\;
\pi_{\text{ref}}
\right)
\right),
\end{align}
where the probability ratio is defined as:
\begin{equation}
r_{\text{ratio}} =
\min
\left(
\frac{\pi_{\theta}(\mathbf{a}_i)}
{\pi_{\theta_{\text{ref}}}(\mathbf{a}_i)},
\;
\text{clip}
\left(
\frac{\pi_{\theta}(\mathbf{a}_i)}
{\pi_{\theta_{\text{ref}}}(\mathbf{a}_i)},
\;
1-\epsilon,
\;
1+\epsilon
\right)
\right).
\end{equation}

This objective encourages the policy to increase the likelihood of action sequences that lead to goal-consistent future states. Importantly, this reinforcement process is performed entirely within the learned latent world model, without requiring additional real-world interaction or action annotations. By iteratively sampling, evaluating, and optimizing in the latent space, the policy can continuously improve and discover more effective action sequence beyond the initial imitation distribution.

\section{Experiment}

\subsection{Experimental Setting}
\subsubsection{Datasets.}
We conduct experiments using a total of three datasets. For the public dataset, we use the datasets RECON \cite{shah2021rapid} and SCAND \cite{karnan2022socially} used by NWM \cite{bar2025navigation} respectively. The RECON dataset is a fisheye camera dataset collected in outdoor environments, which includes a large number of lawn houses, etc. The SCAND dataset is collected in the campus environment, containing a large number of dynamic objects and human interaction scenarios. We selected data from all wheeled robots for approximately three hours. In addition, to validate the practicality of our framework, we collect our own dataset LWM (navigation dataset in Latent World Model) using the robot shown in Fig. \ref{fig:real1}. The dataset includes campus, residential area, and park scenes, covering approximately 6 hours and 600,000 $m^2$. We use wheel encoders and IMUs to obtain the position of the robot, sample one keyframe over 0.2m, and set the maximum predicted frame $K=64$. We use 5\% of the dataset as the test set and the remaining 95\% as the training set. Within the training set, half the data is used to train the world model and imitation learning, while the other half is used to reinforce the world model.

\subsubsection{Baselines.}
\begin{itemize}
    \item NWM \cite{bar2025navigation} is a world model based on image generation for robot navigation tasks. When planning and making decisions, the world model generates different image sequences based on different actions. We follow the approach in the paper to select the action sequence with the minimum LPIPS \cite{zhang2018unreasonable} loss between the goal image and the generated image. The model has already been trained on RECON and SCAND. To align with our evaluation method, we fine tuned NWM-CDiT/XL-1B model for 10 epochs on these two datasets and 30 epochs on LWM.
    \item Dino-WM \cite{zhou2024dino} is a world model based on feature space. All observed images are encoded into features by a Dino-V2 encoder \cite{zhou2024dino}. The features of observation image and actions are fed into the world model to predict the feature of the goal image. During deployment, Dino-WM also used different action sequences for feature prediction, and ultimately selects the action sequence with the smallest error compared to the feature of goal image for execution.
    \item NoMaD \cite{sridhar2024nomad} is a robot navigation method based on diffusion policy \cite{chi2025diffusion}. The current image and the goal image are encoded as feature conditions, and then the noise is guided to transform into an action sequence.
    \item BC is a baseline robot navigation method we designed based on ViNT \cite{shah2023vint} and GNM \cite{shah2023gnm}. It is based on imitation learning. The current and the goal image are input to an image encoder and then decoded into an action sequence.
\end{itemize}
For fair comparison with baseline methods, we deploy three LWM: LWM-S uses Dino-v2-small \cite{oquab2023dinov2} as encoder which has the same 49M parameters as Dino-WM \cite{zhou2024dino}. LWM-B and LWM-B-P uses Croco-v2-base \cite{croco_v2} as encoder which has 217M parameters. LWM-B-P is given the goal prompt in advance as described in \ref{best_practice}. We compare policy model with the same number of parameters. Specific implementation details can be found in the supplementary materials.

\subsubsection{Evaluation Metrics.}
For the evaluation of the world model, we measure prediction accuracy. We use the 64 action sequences in Section \ref{policy_learning} obtained from clustering as candidate trajectories. Following NWM \cite{bar2025navigation}, for each sample in the test dataset, we roll out all candidate action sequences in the world model and select the trajectory whose final state best matches the goal.

\begin{itemize}
\item \textbf{PE and OE}. To quantify prediction error, we compute the Euclidean distance between the predicted goal coordinates and the corresponding ground-truth goal coordinates. Specifically, we report two metrics: Position Error (translation error) and Orientation Error (rotation error), corresponding to the discrepancies in spatial position and attitude, respectively.

\item \textbf{ACC}. In addition, we evaluate the consistency of the world model by measuring its ability to distinguish the correct action sequence from counterfactual alternatives. Specifically, for each sample, we construct a candidate set consisting of the ground-truth action sequence and several counterfactual action sequences from the dataset. Given the current and goal observation, the world model can select an action sequence from candidate set. We report the classification accuracy under different candidate set sizes, including 3 and 5 candidates, denoted by ACC3 and ACC5.

\item \textbf{SR and LPIPS}. To evaluate whether the policy model can enable robots to reach their goals, we deploy them in real-world environments. If the robot reaches within 0.5m of the goal, it is considered successful. To evaluate the accuracy of reaching the goal, when the robot succeeds in reaching the goal, we compare the minimum LPIPS between the robot observation and the goal image.
\end{itemize}

\begin{table}[htb]
\centering
\caption{World Model Prediction and Planning Evaluation. Comparison of prediction errors (PE, OE) and action selection accuracy (ACC3, ACC5). Our method achieve the lowest errors and highest accuracy compared to baseline methods.}
\label{tab:prediction_error}
\resizebox{1.0\linewidth}{!}{
\begin{tabular}{lcccccccccccc}
\toprule
& \multicolumn{4}{c}{RECON} & \multicolumn{4}{c}{SCAND} & \multicolumn{4}{c}{LWM} \\
\cmidrule(lr){2-5} \cmidrule(lr){6-9} \cmidrule(lr){10-13}
Method & PE $\downarrow$ & OE $\downarrow$ & ACC3 $\uparrow$ & ACC5 $\uparrow$  & PE $\downarrow$ & OE $\downarrow$ & ACC3 $\uparrow$ & ACC5 $\uparrow$  & PE $\downarrow$ & OE $\downarrow$ & ACC3 $\uparrow$ & ACC5 $\uparrow$  \\
\midrule
NWM        & 7.222    & 1.279    & 40.97    & 27.23    & 5.534    & 0.623    & 41.66    & 26.04    & 5.710 & 0.884 & 42.43   & 26.26    \\
DINO-WM    & 5.937 & 1.254 & 84.05 & 73.92    & 3.993 & 0.212 & 41.96 & 24.35 & 4.687 & 0.705 & 55.40 & 40.40    \\
LWM-S (ours) & 1.896 & \textbf{0.377} & 85.34 & 79.20   & 2.862 & 0.207 & 67.87 & 47.66  & 3.200 & 0.404 & 66.40    & 51.60    \\
LWM-B (ours) & 1.283 & 0.394 & 91.45 & 87.50   & 2.481 & 0.215 & 58.03 & 39.37  & 3.360 & 0.388 & 61.60  & 46.40    \\
LWM-B-P (ours) & \textbf{1.264} & 0.425 & \textbf{93.10}    & \textbf{88.14}  & \textbf{0.651} & \textbf{0.122} & \textbf{83.93}    & \textbf{70.98}    & \textbf{1.122} & \textbf{0.125} & \textbf{90.60}  & \textbf{83.80}    \\
\bottomrule
\end{tabular}
}
\end{table}

\subsection{Latent World Model}
Compared to other world models, can our framework better predict and plan? We compare our method with several baseline approaches in Table \ref{tab:prediction_error}. Our method achieves the lowest prediction error and the highest action selection accuracy, demonstrating the effectiveness of compatibility modeling. Although Dino-WM also performs prediction in the latent space, it does not use counterfactual action sequences for enhancement. Moreover, the Dino-V2 encoder remains frozen to avoid mode collapse during training, which constrains the adaptability of the representation to the navigation task. As a result, its performance is bounded by the pre-trained visual features and cannot fully exploit task-specific dynamics. Although NWM achieves strong performance in pixel-space prediction, it performs worst when used for planning with predicted images, yielding the highest error and lowest accuracy. This suggests that visually plausible pixel-level generations do not necessarily preserve accurate spatial or geometric consistency, which is critical for robot navigation.

\begin{figure}[htb]
  \centering
  \includegraphics[height=1.9cm]{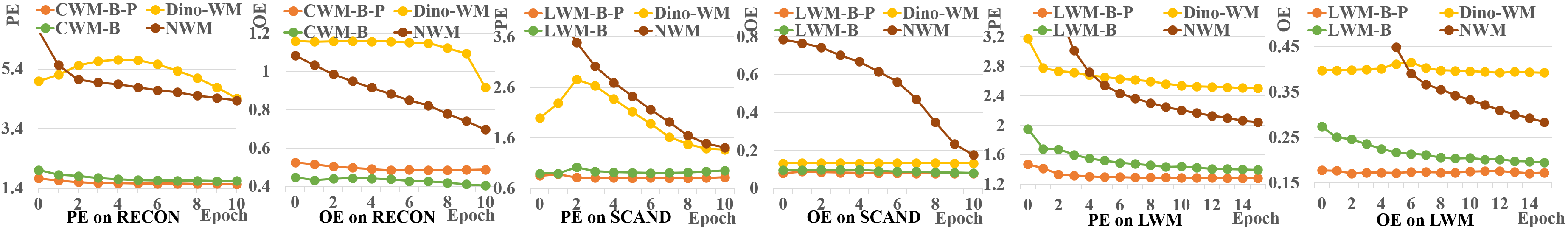}
  \caption{Policy Reinforcement within the World Model. Policies starting from different initializations achieve performance improvements when further optimized via GRPO within LWM.
  }
  \label{fig:grpo}
\end{figure}

\begin{table}[htb]
\centering
\caption{Policy Learning Performance. Evaluation of policies trained under the supervision of different world models demonstrate that LWM provides superior pseudo-labels for policy imitation.}
\label{tab:policy_learning}
\resizebox{0.5\linewidth}{!}{
\begin{tabular}{lcccccc}
\toprule
& \multicolumn{2}{c}{RECON} & \multicolumn{2}{c}{SCAND} & \multicolumn{2}{c}{LWM} \\
\cmidrule(lr){2-3} \cmidrule(lr){4-5} \cmidrule(lr){6-7}
Method & PE $\downarrow$ & OE $\downarrow$ & PE $\downarrow$ & OE $\downarrow $ & PE $\downarrow$ & OE $\downarrow$ \\
\midrule
NWM    & 4.983 & 0.976 & 4.442  & 0.787  & 4.983 & 0.976 \\
DINO-WM    & 4.994 & 1.159 & 1.995    & 0.133    & 3.177 & 0.397 \\
LWM-S (ours) & 2.013    & \textbf{0.448}    & 0.899    & 0.096    & 1.678    & 0.209    \\
LWM-B-P (ours) & \textbf{1.729}  & 0.523 & \textbf{0.853} & \textbf{0.081} & \textbf{1.468}  & \textbf{0.178}    \\
\bottomrule
\end{tabular}
}
\end{table}


\subsection{Learning and reinforcing policy in LWM}

After training the world model, our ultimate objective is to leverage it for policy learning. We evaluate the effectiveness of different world models for policy training in Table~\ref{tab:policy_learning}. For NWM, we use the generated image as the pseudo goal image and the ground truth action sequence as the label. Because the policy model is an autoregressive action generation model, we still report PE and OE as metrics. Our method achieves the lowest error on both metrics, demonstrating that the learned world model provides reliable supervisory signals for training policies directly from unlabeled video data. This capability is particularly valuable in robotics, where action annotations are expensive and scarce.

During the initial policy training stage, we rely on a fixed set of 64 clustered action sequences, which limits the coverage of the action space. To overcome this limitation, we further improve the policy through reinforcement learning within the world model. In practice, policy initialization may come from different sources, so we use the LWM-B-P world model as a unified reward function to evaluate and optimize different initial policies. The reinforcement results are shown in Fig. \ref{fig:grpo}. We observe performance improvements across all policy initializations, demonstrating that our world model can effectively refine policies regardless of their origin. This highlights the generality and practical applicability of our method for scalable policy learning from video data.

\begin{figure}[htb]
  \centering
  \includegraphics[height=3.2cm]{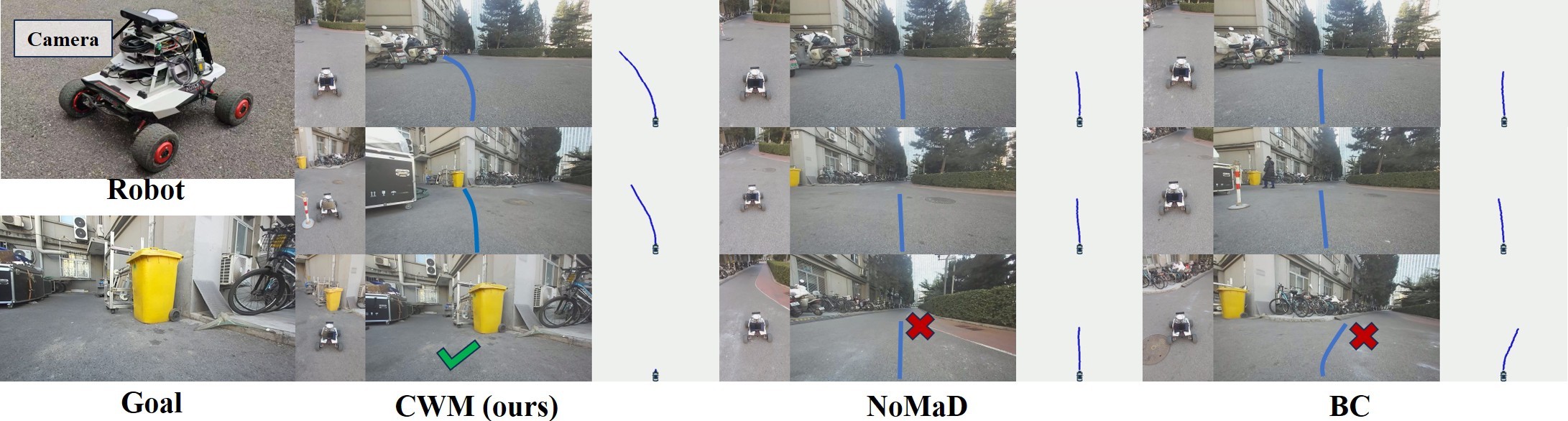}
  \caption{Navigation in real-world environment. LWM can learn policy models within the world model.
  }
  \label{fig:real1}
\end{figure}

\begin{table}[htb]
  \begin{minipage}[c]{0.31\textwidth}
    \centering
    \caption{Real-World Navigation Performance.}
    \label{tab:real_world}
    \resizebox{0.9\linewidth}{!}{
    \begin{tabular}{lcc}
    \toprule
    Method & SR $\uparrow$ & LPIPS $\downarrow$ \\
    \midrule
    NoMaD  & 33.3 & 0.448 \\
    BC    & 33.3 & 0.566 \\
    LWM-B-P (ours) & \textbf{66.7}  & \textbf{0.396} \\
    \bottomrule
    \end{tabular}
    }
  \end{minipage}
  \hfill
  \begin{minipage}[c]{0.68\textwidth}
    \centering

  \includegraphics[height=2.2cm]{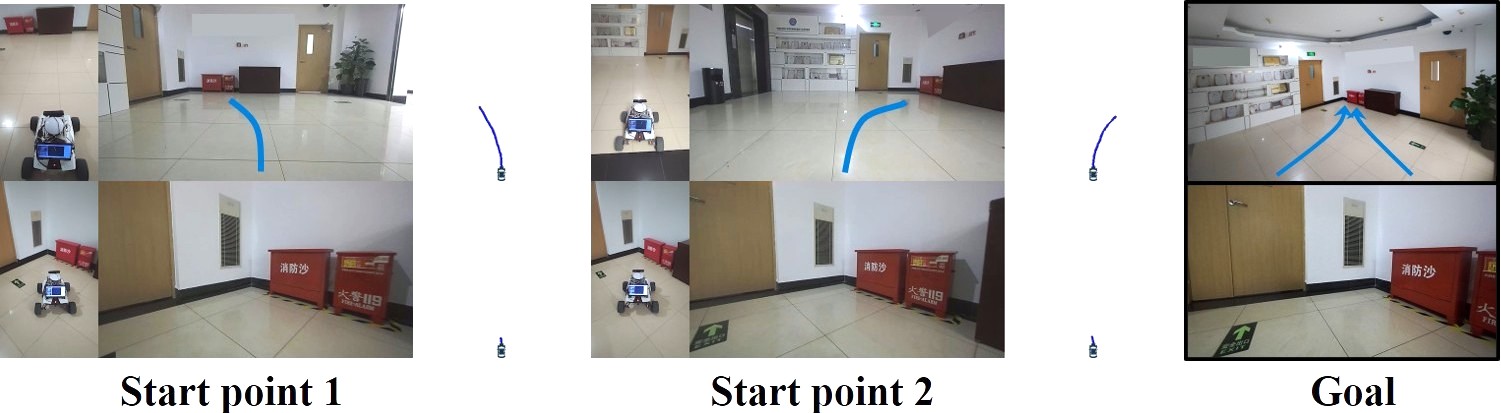}
  \captionof{figure}{LWM demonstrates zero-shot generalization to indoor environments.
  }
  \label{fig:real_indoor}

  \end{minipage}

\end{table}

\subsection{Navigation in the real world}

To validate the effectiveness of our policy trained and reinforced within the world model, we deploy it in real-world robotic environments. We evaluate our method and baseline approaches across six distinct navigation scenarios. The average success rate and the LPIPS are reported in Table~\ref{tab:real_world}. Our method achieves the highest success rate and the lowest LPIPS, demonstrating that our world model has rich imagination and spatial knowledge, even in previously unseen environments. Importantly, our approach does not rely on any action annotations during training. The policy is learned entirely from video data using the world model. As illustrated in Fig. \ref{fig:real1}, our method remains robust even when the goal is located far from the initial position. The policy consistently maintains alignment with the goal state throughout execution. In contrast, imitation learning and diffusion-based policies can perform well in simple scenarios but often fail in more challenging settings, where they tend to lose track of the goal or produce inconsistent trajectories.

Furthermore, we evaluate our method in indoor environments that are completely absent from the training data. As shown in Fig. \ref{fig:real_indoor}, our policy successfully reaches the same goal location from different starting positions. This behavior indicates that the policy has learned meaningful spatial and geometric relationships, rather than merely memorizing action sequences. More real-world robot navigation deployment videos are provided in the supplementary materials.



\begin{table}[htb]
  \begin{minipage}[c]{0.49\textwidth}
    \centering
    \caption{Ablation on Counterfactual Action Sequences. Different action sequences can introduce richer comparisons for supervision and improve performance.}
    \label{tab:counter}
    \resizebox{0.7\linewidth}{!}{
    \begin{tabular}{lcccc}
    \toprule
    Method & PE $\downarrow$ & OE $\downarrow$ & ACC3 $\uparrow$ & ACC5 $\uparrow$ \\
    \midrule
    $ M=1 $  & 2.239 & 0.668 & 34.19 & 23.83 \\
    $ M=2 $  & 1.234 & 0.327 & 61.86 & 44.55 \\
    $ M=4 $  & 1.128 & 0.293 & 64.76 & 52.33 \\
    $ M=8 $  & 1.008 & 0.263 & 65.28 & 46.11 \\
    $ M=16 $  & \textbf{0.890} & \textbf{0.147} & \textbf{70.46}  & \textbf{58.54} \\
    \bottomrule
    \end{tabular}
    }

  \end{minipage}
  \hfill
  \begin{minipage}[c]{0.5\textwidth}
    \centering
    \caption{Ablation on Loss Functions. Our proposed method (MSE with log) effectively captures spatial relationships and subtle action consequences, achieving the best performance.}
    \label{tab:loss}
    \resizebox{1.0\linewidth}{!}{
    \begin{tabular}{lcccc}
    \toprule
    Method & PE $\downarrow$ & OE $\downarrow$ & ACC3 $\uparrow$ & ACC5 $\uparrow$ \\
    \midrule
    Contrastive loss  & 1.298 &\textbf{0.124} & 36.00 & 25.60  \\
    Contrastive and ranking loss    & 1.186 & 0.212 & 85.40 & 80.60\\
    MSE & 1.129  & 0.209 & 86.40 & 79.00\\
    MSE with log (ours) & \textbf{1.122}  & 0.125 & \textbf{90.60} & \textbf{83.80}\\
    \bottomrule
    \end{tabular}
    }

  \end{minipage}

\end{table}

\subsection{Ablation}

We first ablate the effect of modeling interactions among multiple candidate action sequences during training on SCAND dataset using LWM-B-P. Specifically, we vary the number of imagined action sequences evaluated by the world model. As shown in Table \ref{tab:counter}, $M> 1$ indicates that the model uses counterfactual reasoning during training which has better performance. Different action sequences can introduce richer comparisons for supervision and allows the world model to better capture their consequences. For completeness, more extensive results are presented in the supplementary material.

We also study different loss functions on LWM dataset using LWM-B-P. As shown in Table \ref{tab:loss}, using a contrastive loss leads to higher prediction error compared to MSE. This is because contrastive loss only enforces discrete separability between representations, without capturing the magnitude of differences between predicted and target features. We further introduce a ranking loss to encourage better relative ordering among action sequences. While this reduces the prediction error compared to pure contrastive loss, it still underperforms MSE loss. In contrast, our proposed loss achieves the best performance. This indicates that our loss formulation provides more informative supervision, enabling the model to better distinguish subtle differences between action outcomes and focus on harder action sequences.

\section{Conclusion}
We present the compatibility prediction Latent World Model (LWM) for visual navigation. Instead of reconstructing observations, LWM predicts action-conditioned latent feature compatibility via cross-trajectory counterfactual reasoning. This efficient latent space supports policy supervision from unlabeled videos and imagination-driven reinforcement learning, significantly outperforming existing world models in real-world robotic navigation.

\section*{Acknowledgements}
This work was supported in part by the National Natural Science Foundation of China (No. U22B2055, 62273345) and in part by the Beijing Natural Science Foundation (No. L223003).

\clearpage
\section*{Supplementary Material}
\appendix

\section{Details of the Model}

The proposed LWM framework comprises two primary components: the world model and the policy model. In this section, we provide detailed architectural specifications and training hyperparameters for both.

\subsection{World Model}

We evaluate three variants of our world model: LWM-S utilizes DINOv2-small \cite{oquab2023dinov2} as its visual encoder, containing 49M parameters, which matches the capacity of Dino-WM \cite{zhou2024dino}. LWM-B and LWM-B-P employ the CroCo-v2-base \cite{croco_v2} encoder, comprising 217M parameters.To process the actions, we use a standard Multi-Layer Perceptron (MLP) to embed the 3-dimensional action inputs. This MLP features a hidden dimension of 512, an output dimension of 384, and applies a ReLU activation function. The resulting action embeddings are then treated as tokens and augmented with sinusoidal positional encodings. For the generative and discriminative components, we employ standard Transformer decoder architectures. The prediction module consists of 6 decoder layers, 6 attention heads, a feature dimension of 367, and a feed-forward network (FFN) dimension of 2048. The discrimination module follows a similar configuration but is shallower, comprising 4 decoder layers. All networks are trained end-to-end using the Adam optimizer for 50 epochs. We set the batch size to 8 and the peak learning rate to $5 \times 10^{-5}$, accompanied by a linear warmup period of 5 epochs.

\subsection{Policy Model}

We evaluate three distinct policy architectures. To ensure a fair comparison across all experiments, we consistently employ the CroCo-v2-base \cite{croco_v2} visual encoder for all three methods.

\begin{itemize}

\item \textbf{NoMaD \cite{sridhar2024nomad}:} A robot navigation method based on Diffusion Policy \cite{chi2025diffusion}. The current and goal images are encoded and processed by a Transformer encoder featuring 8 layers, 6 attention heads, and a feature dimension of 384. The resulting image features are fed into a 1D U-Net \cite{ronneberger2015u} alongside the diffusion timestep $t$ and noisy actions. The U-Net, which outputs the estimated noise, utilizes three downsampling and three upsampling blocks.

\item \textbf{Behavioral Cloning (BC):} A baseline navigation method inspired by ViNT \cite{shah2023vint} and GNM \cite{shah2023gnm}. Similar to NoMaD, current and goal observations are passed through a Transformer encoder (8 layers, 6 attention heads, feature dimension of 384). The output features are subsequently processed through a pooling layer and projected into action sequences via an MLP.

\item \textbf{LWM Policy (Ours):} Our approach processes the encoded image features as context for an action decoder. This decoder is a standard Transformer decoder consisting of 6 layers, 6 attention heads, and a feature dimension of 384. Analogous to next-token prediction in natural language processing, the model autoregressively generates action tokens, beginning with a special \texttt{[START]} token and concluding upon generating an \texttt{[END]} token.
\end{itemize}

All policy models are trained using the Adam optimizer for 40 epochs with a batch size of 24. We use a learning rate of $1 \times 10^{-4}$ with a linear warmup over the first 4 epochs.

\begin{figure}[htb]
\vspace{-1.0em}
  \centering
  \includegraphics[height=4.5cm]{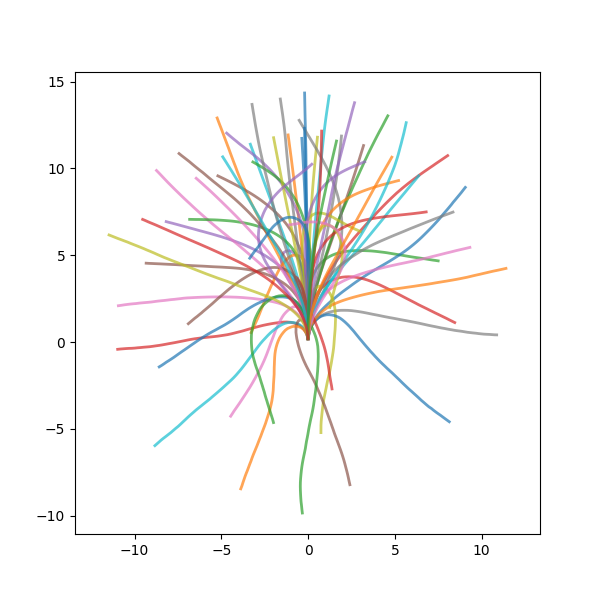}
  \caption{The action sequence centers obtained through k-means clustering.
  }
  \label{fig:kmeans}
  \vspace{-1.0em}
\end{figure}

\section{Ablation of counterfactual action training}

In the experimental section, we presented the ablation results of LWM-B-P on the SCAND \cite{karnan2022socially} dataset. Due to space limitations, we present more ablation experimental results on counterfactual training in LWM dataset here.

\begin{table}[htb]
    \centering
    \caption{Ablation on Counterfactual Action Sequences using LWM-S}
    \label{tab:aba1}
    \resizebox{0.4\linewidth}{!}{
    \begin{tabular}{lcccc}
    \toprule
    Method & PE $\downarrow$ & OE $\downarrow$ & ACC3 $\uparrow$ & ACC5 $\uparrow$ \\
    \midrule
    $ M=1 $  & 2.800 & 0.968 & 43.96 & 30.63 \\
    $ M=2 $  & 1.143 & 0.210 & \textbf{89.36} & 80.18 \\
    $ M=4 $  & \textbf{1.100} & \textbf{0.197} & 88.46 & \textbf{80.36} \\
    $ M=8 $  & 1.138 & 0.210 & 86.84 & 77.83 \\
    $ M=16 $  & 1.142 & 0.205 & 87.20 & 77.83 \\
    \bottomrule
    \end{tabular}
    }

\end{table}

\begin{table}[htb]
    \centering
    \caption{Ablation on Counterfactual Action Sequences using LWM-B-P}
    \label{tab:aba2}
    \resizebox{0.4\linewidth}{!}{
    \begin{tabular}{lcccc}
    \toprule
    Method & PE $\downarrow$ & OE $\downarrow$ & ACC3 $\uparrow$ & ACC5 $\uparrow$ \\
    \midrule
    $ M=1 $  & 5.205 & 1.083 & 32.97 & 20.72 \\
    $ M=2 $  & 3.586 & 0.467 & \textbf{60.36} & 46.18 \\
    $ M=4 $  & \textbf{3.307} & 0.412 & 62.88 & \textbf{49.54} \\
    $ M=8 $  & 3.427 & \textbf{0.408} & 61.26 & 47.20 \\
    \bottomrule
    \end{tabular}
    }

\end{table}

Figure \ref{fig:kmeans} visualizes the action sequence centers obtained through k-means clustering. The quantitative results, presented in Tables \ref{tab:aba1} and \ref{tab:aba2}, demonstrate that the absence of counterfactual training severely degrades model performance. Conversely, the introduction of counterfactual training leads to progressive performance gains that eventually saturate. This phenomenon indicates that our explicitly defined compatibility labels act as a strong initialization rather than an absolute upper bound. The model leverages this supervisory signal to inherently refine the distance metric of the feature space. Consequently, this confirms that a well-designed inductive bias is instrumental in boosting the model's performance.

\begin{figure}[htb]
\vspace{-1.0em}
  \centering
  \includegraphics[height=5.4cm]{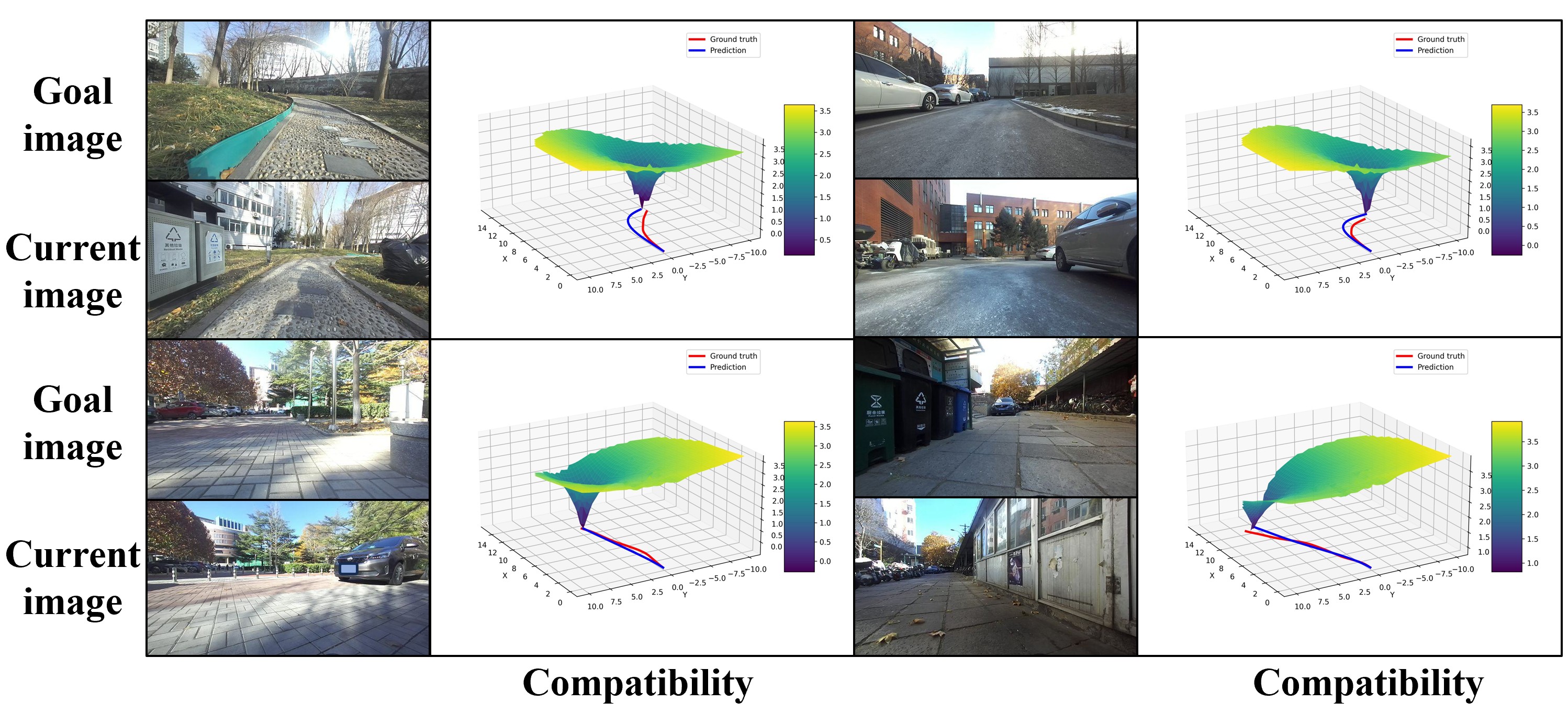}
  \caption{The compatibility surface predicted by LWM.
  }
  \label{fig:Compatibility_vis}
  \vspace{-1.0em}
\end{figure}

\section{Navigation in real world}

To investigate whether our world model can effectively predict compatibility in real-world environments, we visualize its predictions on the LWM test dataset, as illustrated in Fig. \ref{fig:Compatibility_vis}.  The visualization plots the distance function, where lower values along the z-axis denote higher compatibility. Notably, the world model yields a smooth surface, with the maximum compatibility correctly peaking near the ground-truth value. This topological smoothness is highly advantageous for policy learning and reinforcement within the world model. Specifically, as the policy model explores the action space around the ground truth, this smooth landscape provides stable and informative gradients, effectively guiding the optimization process toward the optimal behavior. We provide the video of robots deployed in the real world in another file.

\section{Limitations and Future Work}
While our imagination-based world model achieves state-of-the-art performance in navigation tasks, we acknowledge certain boundary conditions. A core inductive bias of this approach is the assumption that spatial proximity strongly correlates with visual similarity. While this provides a highly efficient supervisory signal in standard scenarios, marginal spatial displacements in densely cluttered environments with severe occlusions can cause abrupt visual alterations. Under such partial observability, visual similarity may temporarily decouple from spatial proximity, making compatibility predictions less discriminative. To address these extreme edge cases, future work could explore integrating 3D-aware latent representations or long-term temporal memory, ensuring robust compatibility scores even during sudden, occlusion-induced visual shifts.

\bibliographystyle{unsrtnat}
\bibliography{references}
\end{document}